# Fine-tuning and Utilization Methods of Domain-specific LLMs

Cheonsu Jeong[1]

Dr. Jeong is Principal Consultant & the Technical Leader for AI Automation at SAMSUNG SDS

## Abstract

Recent releases of pre-trained Large Language Models (LLMs) have gained considerable traction, yet research on fine-tuning and employing domain-specific LLMs remains scarce. This study investigates approaches for fine-tuning and leveraging domain-specific LLMs, highlighting trends in LLMs, foundational models, and methods for domain-specific pre-training. Focusing on the financial sector, it details dataset selection, preprocessing, model choice, and considerations crucial for LLM fine-tuning in finance. Addressing the unique characteristics of financial data, the study explores the construction of domain-specific vocabularies and considerations for security and regulatory compliance.

In the practical application of LLM fine-tuning, the study outlines the procedure and implementation for generating domain-specific LLMs in finance. Various financial cases, including stock price prediction, sentiment analysis of financial news, automated document processing, research, information extraction, and customer service enhancement, are exemplified. The study explores the potential of LLMs in the financial domain, identifies limitations, and proposes directions for improvement, contributing valuable insights for future research. Ultimately, it advances natural language processing technology in business, suggesting proactive LLM utilization in financial services across industries.

---

[1] Corresponding and first author: jeongcsmon@gmail.com

# 1. Introduction

In recent years, advancements in artificial intelligence and deep learning technologies have led to remarkable achievements, particularly in the field of natural language processing. Large Language Models (LLMs), in particular, have reached human-level language generation and comprehension capabilities with their rich context and language understanding. Consequently, various industries are actively considering or implementing the adoption of LLMs. According to McKinsey, generative AI, including LLMs, is anticipated to contribute to productivity enhancement in the banking industry, specifically in areas such as marketing/sales, customer support/management, programming, and regulatory compliance. The potential value creation by generative AI in the global banking industry is projected to range from $200 billion to $340 billion, corresponding to 2.8% to 4.7% of the industry's total revenue. Financial institutions are increasingly leveraging LLMs to support and automate employee tasks internally, as well as to collect and analyze natural language-based information for strategic decision-making (Financial Focus, 2023). Even in the finance sector, traditionally slow in adopting new technologies, there is a growing expansion of the use of generative AI, specifically LLMs.

Therefore, this study aims to explore fine-tuning and utilization cases of LLMs, particularly those applied in the financial domain. The financial sector is characterized by rapidly changing market environments, diverse financial events, and vast amounts of financial data. To address these challenges, rapid and accurate information processing and decision-making capabilities are essential. Recent research suggests that LLMs can effectively address these challenges, and the market has witnessed active releases of general LLMs. However, in-depth research that comprehensively understands and addresses the various issues arising during the construction and utilization of LLMs tailored for specific tasks is still lacking.

This study seeks to propose specific methods for applying domain-specific LLMs, with a focus on the financial industry. The financial sector possesses specific characteristics, such as trust, consumer protection regulations, and inclusive finance, distinguishing it from other industries (Financial Focus, 2023). The financial domain, being sensitive to technological innovation while prioritizing stability and accuracy, underscores the importance of developing and utilizing domain-specific LLMs to enhance the efficiency and competitiveness of financial businesses.

This research, within this context, is expected to provide a deep understanding of the core technology and potential applications of domain-specific LLMs in the financial industry, delivering practical value to financial institutions and research organizations. The study specifically addresses LLM fine-tuning and application cases in the financial domain. However, the financial sector is extensive, comprising various subfields, and further research is needed to cover specific areas not addressed in this study. Additionally, the interpretation of research results will consider the limitations and constraints of language models. The primary aim of this research is to explore effective methodologies for fine-tuning domain-specific LLMs in the financial sector and to study real-world use cases in financial tasks. It is anticipated that language models reflecting expertise in the financial domain will demonstrate enhanced performance, contributing to increased productivity and decision support in financial tasks.

This paper is organized as follows: Chapter 2 examines LLMs and fine-tuning, Chapter 3 provides detailed explanations of approaches to create domain-specific LLMs tailored for the financial sector. Chapter 4 explores various perspectives on LLM creation methods and their application in the financial sector, and Chapter 5 concludes the research and discusses future directions.

## 2. Related Work

For this study, recent major research papers, journals, articles, and books related to generative AI and LLM were investigated. In this chapter, an overview of LLMs, generative AI, and foundation models is discussed, followed by an in-depth exploration of LLM fine-tuning. The application areas of language models in the financial domain, as covered in this paper, are also outlined.

### 2.1. Overview of Large Language Models (LLMs)

In recent years, LLMs have witnessed distinctive advancements in the field of Natural Language Processing (NLP). Models such as BERT (Bidirectional Encoder Representations from Transformers) and GPT (Generative Pre-trained Transformer), based on the Transformer architecture, have demonstrated powerful representations learned through pre-training on massive text data, making them applicable to various NLP tasks. These models showcase excellent performance in understanding text context, diverse grammatical structures, and semantic relationships. Generative AI, as a form of artificial intelligence, utilizes extensively trained data models to generate new content, including text, images, audio, and video (Jeong C.S., 2023d). In the NLP domain, advancements in Natural Language Understanding (NLU) technologies have enabled complex dialogue processing through the utilization of Context models and Transformer language models (Jeong, 2023a). Recently, the integration of chatbots with other solutions such as Robotic Process Automation (RPA) and Optical Character Recognition (OCR) has been observed to directly enhance efficiency in various tasks (Jeong C.S., Jeong J.H., 2020). In November 2022, OpenAI introduced the ChatGPT model, an AI chatbot trained through a reinforcement learning process involving feedback from a group of human experts, enhancing the conversational capabilities of the GPT-3.5 model. ChatGPT garnered significant attention, surpassing 100 million monthly users within two months of its release (Jeong C.S., 2023b). In open-domain chatbot conversations like ChatGPT, an interface with LLM models occurs through prompts, and caution must be exercised when applying parameters between applications to avoid handling sensitive personal information. Although laws exist to prevent developers from collecting and using user data without consent, users find it challenging to understand how much data developers collect and where this data is stored in real-life situations (Jeong, J. H., and Jeong, C. S., 2022). Additionally, in conversation with ChatGPT, the completeness of responses depends on how detailed the question or request prompt is. Therefore, investigating prompt engineering, which involves finding combinations of prompt input values from LLMs, is crucial (Jeong C.S., 2023c). Specifically, in the financial sector, the importance of providing up-to-date information through customer response chatbots is emphasized, and limitations in the information capacity and hallucination issues of LLMs are identified as challenges. Approaches to address these challenges include fine-tuning with new data and directly inserting information into prompt contexts. However, fine-tuning incurs significant costs, and including all information in prompts is practically challenging. As an alternative, the Retrieval-Augmented Generation (RAG) model has been proposed, as illustrated in Figure 1. This model involves storing information in vector databases and searching for the required information to be presented to the LLM (Jeong C.S., 2023e).

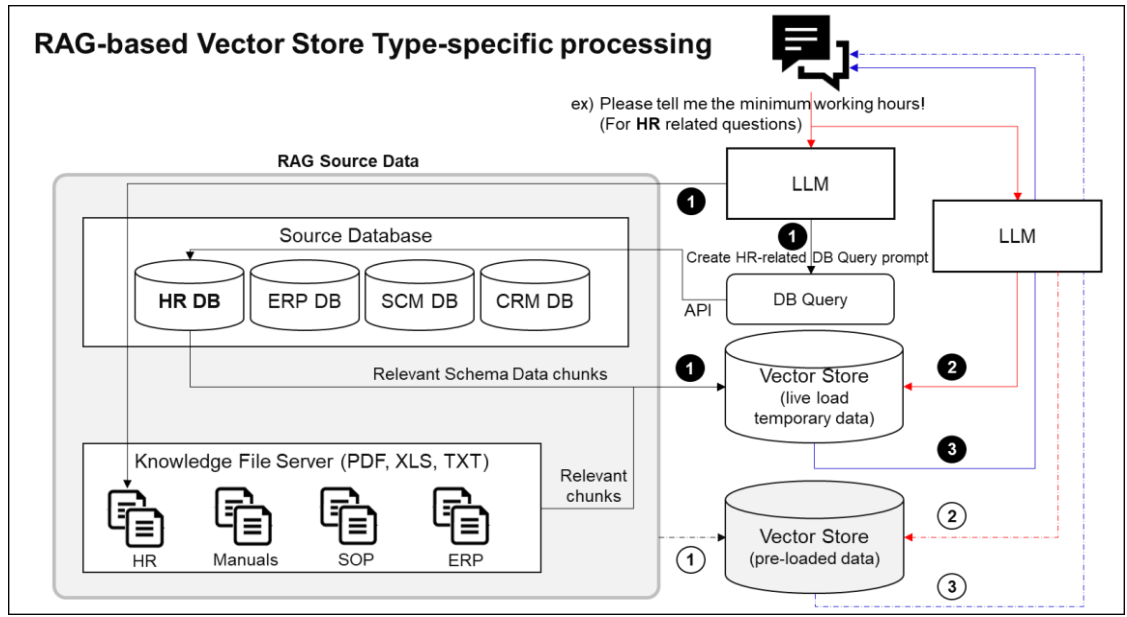

<Figure 1> RAG-based Vector Store configuration type and processing procedure

Additionally, LLMs and generative AI, as depicted in Figure 2, are positioned within AI's deep learning, allowing for the utilization of LLMs based on deep learning to provide generative AI services (Mayank, S., 2023; Jeong C.S., 2023e).

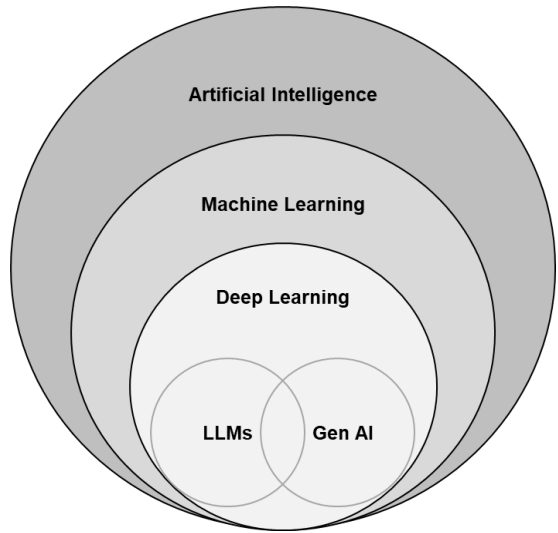

<Figure 2> LLM & Generative AI Relation Diagram

2.1.1. Trends in Generative AI

In recent years, as of 2023, the development of super-sized AI technologies has been rapidly advancing. Table 1 illustrates the current status of LLMs and generative AI services introduced in 2023.

<Table 1> Status of generative AI model launches in 2023

| Country | Company | Foundation Model | Parameters | Source | Release Date | Service |
|---|---|---|---|---|---|---|
| USA | OpenAI / MS | GPT-4 Turbo | 1.75T | Closed | 2023.11 | ChatGPT, GPTs / MS Bing AI, MS Copilot, MS 365 Copilot |
| | Google | Gemini | Closed | Closed | 2023.12 | Bard |
| | META | LLaMA2 | 7~65B | Open | 2023.07 | Code Llama |
| | Stanford Univ. | Alpaca | 7B | Open | 2023.03 | LLaMA 7B Fine-tuning Model |
| | Nomic | GPT4All v2 | 3~13B | Open | 2023.04 | LLaMA 7B Fine-tuning Model |
| | Hugging Face | BLOOM | 176B | Open | 2022.07 | - |
| South Korea | Naver | HyperClovaX | Closed | Closed | 2023.08 | Polaris Office AI, Lewis, etc. |
| | LG | EXAONE2.0 | 300B | Self-utilization | 2023.07 | AI artist Tilda, etc. |
| | NC Soft | VARCO | Closed | Closed | 2023.08 | VARCO Art/Text/Human/Studio |
| | SKT | A. Enterprise | Closed | Self-utilization | 2023.08 | Document summary & creation, etc. |
| | KT | MI:DEUM2 | 200B | Self-utilization | 2023.10 | GiGA Genie, AICC, AI care service |
| | SAMSUNG | Samsung Gauss | Closed | Self-utilization | 2023.11 | Gauss Language/Code/Image |
| China | Huawei | PanGu 3.0 | 100B | Open | 2023.07 | Pangu-Weather, etc. |
| | Baidu | Ernie 3.5 | 130B | Self-utilization | 2023.06 | Ernie Bot 3.5 |
| | Alibaba | Tongyi Qianwen | 10T | Self-utilization Open | 2023.04 | DingTalk, Tmall Genie Open source service: ModelScope |
| UAE | TII | Falcon | 7~180B | Open | 2023.09 | - |

In March, OpenAI unveiled GPT-4, and in May, Google introduced PaLM2. While the parameter count was reduced compared to previous models, they achieved higher performance by training on approximately five times more tokens (text data). Additionally, in November, Samsung publicly revealed Samsung Gauss, stating plans to gradually incorporate Samsung Gauss into products such as the Galaxy S24 to be released in the future. Noteworthy companies like Kakao are either in the process of constructing or considering developing their own generative AI. There is a growing interest in open-source LLMs such as Meta's LLaMA (Large Language Model Attention) and Falcon, with a focus on training volume rather than model size (Jeong C.S., 2023e).

2.1.2. Foundation Model and LLM Pre-training

Generative AI models, depending on the type of output they generate, employ language models, image models, video models, etc. However, currently, multi-modal models, capable of simultaneously learning from both images and text, are rapidly evolving, establishing themselves as foundation models (Jeong C.S., 2023e). The data for the foundation model, which encompasses text, images, audio, structured data, 3D signals, etc., is utilized without distinction during training. These models, including human creativity and reasoning abilities, represent a shift in the AI paradigm and are referred to as foundation models. Massive amounts of data, obtained through unsupervised learning, are deployed and fine-tuned for downstream tasks such as fine-tuning or in-context learning, completing the foundation model to suit the user's desired objectives (Bommasani, R., et al., 2021).

Among foundation models, Large Language Models (LLMs) as language models are pre-trained language models (PLMs) that utilize vast amounts of general knowledge, such as text data from sources like Wikipedia, collected through self-supervised or semi-supervised learning. The foundational process involves gathering the training dataset, which serves as the learning resource for the LLM. The data can be sourced from various outlets, including books, websites, articles, and public datasets. To develop a proficient LLM, a text dataset from pre-trained material is used. The sources of pre-trained corpora can be broadly classified into two types: general data and specialized data. General data, such as web pages, books, and conversational texts, is extensive, diverse, and readily accessible, making it predominantly utilized in most LLMs. It enhances the language modeling and generalization capabilities of LLMs. Additionally, there are studies expanding the pre-trained corpus to more specialized datasets, such as multilingual data, scientific data, and code, to give LLMs specific task-solving capabilities (Taylor, R., et al., 2022; Chowdhery, A., et al., 2022; Nijkamp, E., et al., 2022).

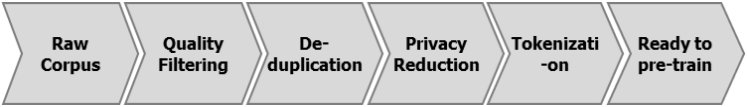

<Figure 3> LLM Pre-training Process

Figure 3 illustrates the typical data collection and pre-training process for LLMs (Zhao, W. X., et al., 2023; Jeong C.S., 2023e). Model training involves using supervised learning on pre-processed text data. Given the substantial sizes of both the model and the data, model parallelism is employed to facilitate training by requiring substantial computational power. Training a large-scale language model from scratch demands significant investment; therefore, a more economical alternative is to fine-tune existing language models for specific use cases (Jeong C.S., 2023e).

2.1.3. Domain-Specific LLM Pre-training

When aiming to create an LLM specialized in a particular domain, the process involves selecting the target domain, collecting relevant data, injecting specific domain data information into a general LLM, and then training a conversational model for the selected domain (Adaptive Pre-training).

Several criteria guide domain selection, including the likelihood of a domain being already learned, the uniqueness of a niche or recently updated domain, the presence of at least 10,000 characters of text data describing the domain, and the verification of learning status for domains with formal Wikipedia or Namuwiki information.

## 2.2. LLM Fine-Tuning

While Large Language Models (LLMs) are trained on extensive datasets and possess general knowledge, they may not perform optimally in specific tasks without fine-tuning. Therefore, as depicted in Figure 4, the model's performance is enhanced through a fine-tuning process. For models of a smaller scale, such as BERT (450M) or RoBERTa (1.3G), full fine-tuning was traditionally conducted. However, with the introduction of LLaMA, fine-tuning these large models in their entirety became computationally challenging. Approaches like LoRA, which involves fixing the weights of existing pre-trained layers and only training the weights of new layers, were employed. Recently, it has been validated that there is not a significant difference in actual performance, leading to the adoption of the Parameter-efficient Fine-tuning (PEFT) method. PEFT involves adding a small number of new parameters to the pre-trained LLM, fine-tuning only the added parameters, and achieving better performance at a lower cost (Raschka, S., 2023).

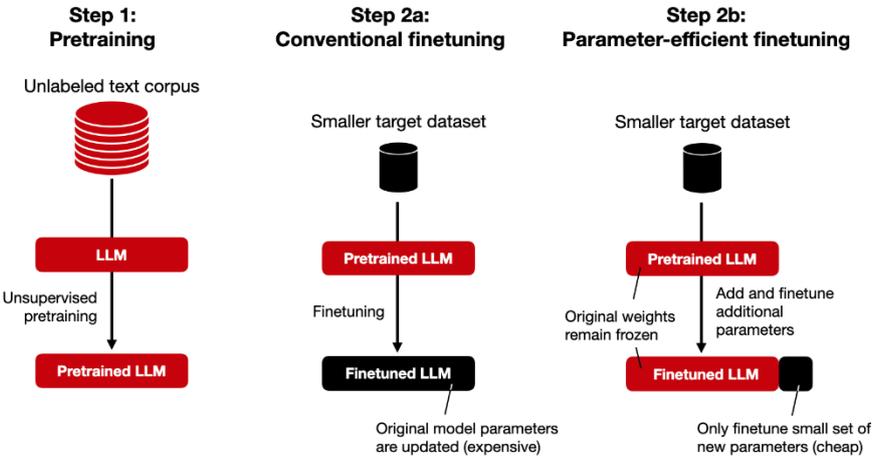

<Figure 4> Fine-tuning of LLM

LoRA, a prominent approach, allows additional learning for specific weight matrices within Large Language Models (LLMs). Within each Transformer Layer, it determines the target for fine-tuning by selecting specific weight matrices. These matrices are replicated in their original form, and fine-tuning proceeds with allowing learning for the chosen weight matrices. The weight matrices undergo Low-Rank Matrix Decomposition, forming two matrices, as illustrated in Figure 5.

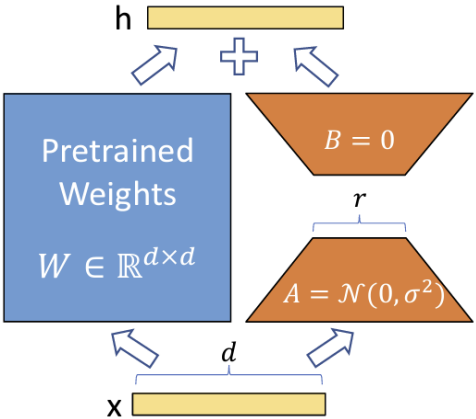

<Figure 4> LoRA Re-Parameterization

The A matrix is subject to Random Gaussian initialization, while the B matrix is initialized to Zero at the beginning of the learning process (Edward, H., et al., 2021). This methodology ensures that fine-tuning is selectively applied to essential weight matrices, optimizing the LLM for specific tasks.

Given that pre-training data significantly influences the performance of Large Language Models (LLMs), the demand for high-quality data for model pre-training is even more crucial for LLMs compared to small language models. The model's capacity relies heavily on the corpus data collected for pre-training and the pre-training processing methods (Jeong, C.S., 2023e). LLMs, serving as foundational models that leverage deep learning for Natural Language Processing (NLP) and Natural Language Generation (NLG) tasks, undergo pre-training on extensive datasets to learn the complexity and connectivity of language. Various technologies, including fine-tuning, in-context learning, zero/one/few-shot learning, are employed to facilitate this learning process (Dilmegani, C., 2023).

After pre-training, the model is evaluated on a test dataset that was not used for training to measure its performance. Based on the evaluation results, hyperparameters may be adjusted, architecture modified, or additional training on new data may be performed to fine-tune the model and enhance its performance (Jeong, C.S., 2023e). The resulting Fine-tuned Language Model (FLM) is then applied in various domains, serving as a specialized small language model (SLM).

### 2.2.1. Recent Advances in Fine-Tuning

The recent advancements in fine-tuning, particularly the Parameter-efficient Fine-Tuning (PEFT) method, can be categorized into Prompt Modification, Adapter Methods, and Parameterization. Prompt Modification includes Hard Prompt Tuning, Soft Prompt Tuning, and Prefix-tuning. Adapter Methods, such as LLaMA-Adapter, aim to minimize the side effects of fine-tuning on the entire model by introducing modularized parameters through adapters. In this process, a single adapter module, centered around the Bottleneck Layer, performs linear transformations of Down-projection and Up-projection, while the pre-trained LLM is frozen during fine-tuning.

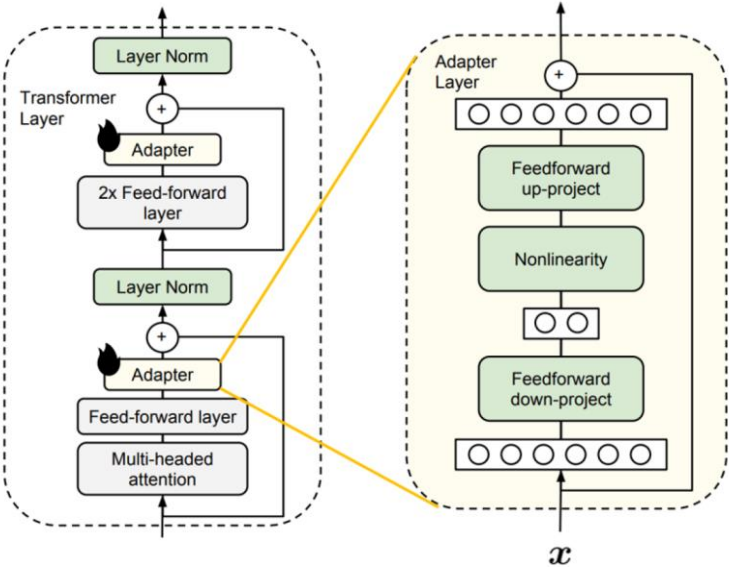

<Figure 6> Adapter methods

Notably, QLoRA, an improved version of LoRA, has been introduced. QLoRA provides a method to PEFT LLM using QLoRA, allowing testing of LLM models even on personal computers. Unlike LoRA, which concatenates additional data by keeping the base model's network intact, QLoRA introduces 16-bit network nodes quantized to 4 bits and employs a paging mechanism for swapping binary data to handle large models with limited memory. Although there is some information loss in reducing from 16 bits to 4 bits, it is considered acceptable (Dettmers, T., et al., 2023).

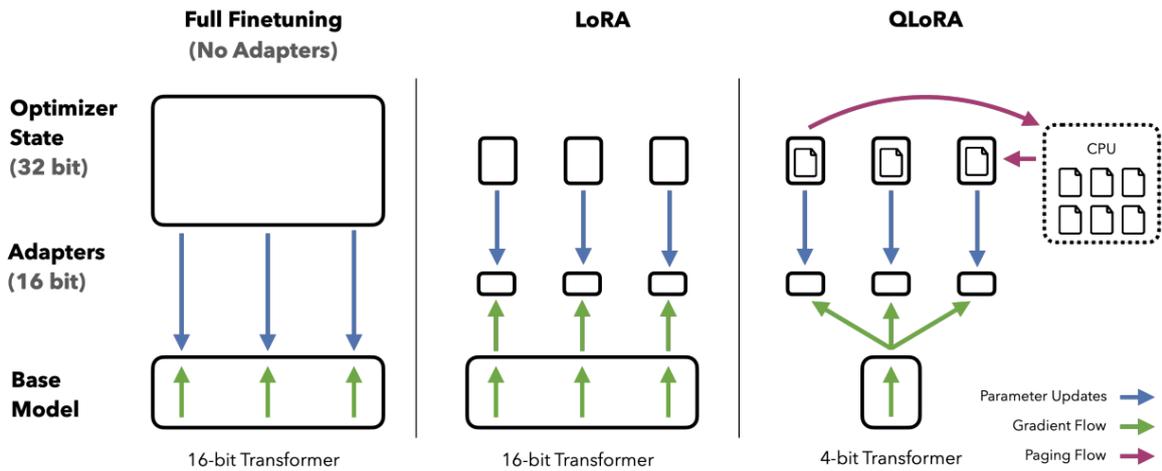

<Figure 7> QLoRA improves over LoRA

Another prevalent fine-tuning approach is Instruction Tuning, where the task is transferred to natural language. Unlike Vanilla-LLM, which simply completes the next text, Instruction Tuning LLM completes the next text based on user instructions. This involves training the model to understand and execute task instructions, enabling it to infer new tasks according to the provided instructions. Alpaca LLM, for instance, is a model based on LLaMA 7B, fine-tuned through instruction tuning.

## 3. Methods

In this section, we present a systematic approach for generating finance-specialized LLM. The fine-tuning process follows the procedure outlined in Figure 8. The data collection and preprocessing stages involve selecting finance-specific datasets and implementing effective preprocessing methods. In the model selection and fine-tuning steps, a suitable pre-trained LLM, known as Pre-trained Language Model (PLM), is chosen, and hyperparameters are tuned accordingly. Considerations for fine-tuning in the context of finance include addressing financial data characteristics, domain-specific vocabulary, and fine-tuning algorithms.

Next, in the configuration of LLMs for the finance domain, we introduce approaches for both training models from scratch and tuning existing models. Lastly, in the evaluation criteria and metrics section, we guide the evaluation of the model using both quantitative and qualitative performance metrics. Through this, we aim to comprehensively outline the creation of finance-specialized LLMs.

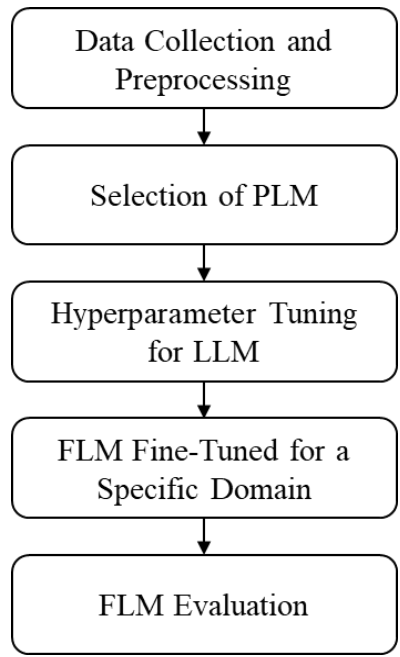

<Figure 8> Overview of Fine-tuning Procedure

## 3.1. Data Collection and Preprocessing

### 3.1.1. Selection of Finance-Specific Datasets

Firstly, selecting datasets specialized in the finance domain is crucial. To achieve this, various sources such as financial research reports, financial news, and market transaction records will be utilized. Finance datasets can exist in diverse forms, including financial-related texts, codes, images, audio, etc. Consider the following factors when selecting finance-specific datasets:

1) **Type of Data:** Finance datasets can take the form of text, code, images, audio, etc. Each type has distinct characteristics, so choose data that aligns with the intended purpose.

2) **Quantity of Data:** The volume of data significantly influences model performance. Generally, larger datasets lead to improved model performance.

3) **Quality of Data:** Data quality impacts model accuracy. Ensure that the data is free from errors or biases, as these can degrade model performance.

Two primary methods can be employed to select finance-specific datasets:

1) **Using Existing Datasets:** Utilize publicly available datasets in the finance domain. Examples include:

. Financial News Dataset: Collected news articles for sentiment analysis related to financial markets.

. Financial Document Dataset: Compilation of financial reports, contracts, regulations, etc., for understanding financial products and services.

. Financial Code Dataset: Collection of financial-related software code for understanding and automating financial systems.

2) **Creating a Custom Dataset:** Building a dataset from scratch provides the advantage of tailoring it to specific needs. Considerations include:

. Data Collection Method: Choose methods such as web scraping, database querying, or sensor data, depending on the intended use.

. Data Cleaning Method: Remove errors or biases during data cleaning.

These datasets can enhance sentiment analysis of financial news or recommend financial products more effectively based on customer characteristics.

### 3.1.2. Data Preprocessing Methods

The collected financial data is likely to be irregular and complex. Therefore, preprocessing the data to a format suitable for model training is necessary. Techniques such as text normalization, tokenization, stop-word removal, and morphological analysis are applied to refine the data.

In the finance domain, commonly used data preprocessing methods include:

1) **Text Data Preprocessing:**

. Character Normalization: Removing special characters, emojis, and extra spaces.

. Word Tokenization: Breaking sentences into individual words.

. Stop-word Removal: Eliminating meaningless words.

. Building Vocabulary: Storing word meanings and features in a dictionary.

2) **Code Data Preprocessing:**

. Code Cleanup: Standardizing code format and correcting errors.

. Code Structure Analysis: Understanding code structure and dividing it into meaningful units.

. Code Meaning Analysis: Comprehending code meaning and extracting relevant

information.

3) **Image Data Preprocessing:**

   . Image Normalization: Adjusting image size, brightness, contrast, etc.

   . Feature Extraction: Extracting features from images.

   . Image Classification: Categorizing images.

These preprocessing steps ensure that the finance model comprehends the nuances of financial products more accurately.

## 3.2. Model Selection and LLM Fine-Tuning Process

In this section, the essential steps for selecting a model and fine-tuning a Language Model (LLM) for the finance domain are covered. Firstly, an appropriate pre-trained LLM model is chosen, and then the model undergoes a fine-tuning process to specialize it for the finance domain. This approach allows the construction of a language model sensitive to the nuances of the finance industry.

### 3.2.1. Selection of Pre-trained LLM Model

To develop a finance-specific LLM, the choice of a pre-trained language model is crucial. The selected model should have language comprehension abilities that extend to the specific terminologies and contexts of the finance domain. Consider the following criteria when choosing a pre-trained LLM model:

1) Model Size: The size of the model impacts its performance. Generally, larger models tend to deliver better results.

2) Model Purpose: Different models may be more suitable depending on their intended use. For tasks like sentiment analysis of financial news, a model specialized in natural language processing is preferable.

3) Model Availability: If a model is not publicly available, training the model from scratch becomes necessary. This process can consume considerable time and resources.

When selecting a pre-trained LLM model, consider the following elements:

1) Performance and Applicability: Assess the model's performance and suitability for finance applications.

2) Training Data Characteristics: Ensure that the model's training data aligns with the characteristics of financial data. The model selection should align with both performance metrics and the specific requirements of the finance domain. It's essential to verify that the chosen model understands the nuances of financial terms and contexts.

Once a pre-trained model is selected, the next step involves fine-tuning it to adapt to the intricacies of the finance domain. The fine-tuning process refines the model's understanding of finance-related language patterns, enhancing its applicability to tasks within this specific industry.

Several pre-trained Language Models (LLMs) are available for applications in the finance domain. Here are some notable models:

1) GPT-4: GPT-4 is a large-scale language model developed by OpenAI. It has 1.75 trillion parameters, making it one of the most powerful language models in the world. GPT-4 can be used in a variety of fields, including natural language processing, machine translation, and code generation. It has been shown to be capable of generating human-quality text, translating languages accurately, and writing different kinds of creative content.

2) BERT: BERT is a large-scale language model developed by Google AI. It has 100 million parameters, making it smaller than GPT-4 but still very powerful. BERT can be used in a variety of fields, including natural language processing, question answering, and sentiment analysis. It has been shown to be very good at understanding the meaning of text and answering questions about it.

3) LLaMA2: LLaMA2 is a large-scale language model developed by Meta. It has 7 to 65 billion parameters, making it a smaller model than GPT-4 but still very powerful. LLaMA2 is unique in that it is released as open source, meaning that anyone can use it for research or commercial purposes. This has made it a popular choice for researchers and developers who want to build their own language models.

4) Falcon: Falcon is a large-scale language model developed by TII in the UAE. It has 7 to 180 billion parameters, making it a very powerful model. Falcon is also released as open source, making it a popular choice for researchers and developers.

5) BloombergGPT: BloombergGPT is a finance-specific large language model developed by

Bloomberg. It has 10 billion parameters, making it smaller than some of the other models on this list but still very powerful. BloombergGPT is trained on a dataset of financial news and can be used to perform a variety of tasks in the financial field, such as sentiment analysis of financial news and recommendation of financial products.

6) FinBERT: FinBERT is a variant of BERT that is specifically designed for the financial field. It is pre-trained with a dataset of financial data and can be used to perform natural language processing tasks in the financial field, such as named entity recognition and relationship extraction.

When selecting a pre-trained LLM model for the finance domain, consider factors such as the model's size, purpose, and availability. The choice between GPT-4, BERT, and recently introduced finance-specialized models like BloombergGPT and FinBERT depends on the specific requirements and goals of the intended applications.

Additionally, models like LLaMA2 and Falcon, which are open-source and versatile, can be considered based on their availability and suitability for the finance domain. The ultimate decision should align with the intended use case and the model's ability to capture the intricacies of financial language and context.

### 3.2.2. Hyperparameter Tuning for LLM

Hyperparameter tuning is a crucial step in maximizing the performance of a model. Adjusting key hyperparameters appropriately during the construction of a domain-specific LLM enhances the model's learning and generalization abilities. Here are some essential hyperparameters to consider:

1) Learning Rate: Determines how much the model's weights should be updated. Set an appropriate initial learning rate to control the convergence speed and prevent issues like divergence or insufficient convergence (Example: Set the initial learning rate to 0.0001).

2) Batch Size: Determines the amount of data processed by the model in each iteration. Choose a batch size that balances the trade-off between allowing more updates and controlling memory usage (Example: Set the batch size to 32 or 64).

3) Number of Epochs: Represents how many times the entire dataset is passed through the model. Avoid too few epochs, which may result in insufficient learning, or too many epochs, which could increase the risk of overfitting (Example: Set the number of epochs to 10 or 20).

4) Dropout Rate: Dropout randomly excludes some neurons during training to prevent the model from relying too heavily on specific patterns. Choose an appropriate dropout rate to enhance the model's generalization ability (Example: Set the dropout rate to 0.2).

5) Other Model-Specific Hyperparameters: Different models may have specialized hyperparameters. For instance, in Transformer-based models, adjusting the number of attention heads or layers can impact model complexity and performance (Example: Adjust the number of attention heads, Transformer layers, etc.).

When tuning hyperparameters, it is common to conduct experiments to find the optimal combination. Continuously evaluate the performance changes resulting from different settings and iteratively refine the hyperparameters to achieve the best learning configuration. This iterative process allows for the discovery of optimal hyperparameter values that contribute to the model's effectiveness in the finance domain.

### 3.2.3. Fine-Tuning Execution Environment Setup

Setting up the fine-tuning execution environment is a critical step for optimizing model training and achieving high performance. To fine-tune a model, it's essential to configure an appropriate execution environment. This involves utilizing high-performance computing resources, including GPU acceleration, to enhance training speed and establishing an environment for real-time monitoring of the model's performance. By considering various factors and configuring the execution environment accordingly, the efficiency and performance of the model can be improved.

1) Hardware Selection: Choose high-performance hardware for fine-tuning large datasets and complex models. Utilize GPU (Graphics Processing Unit) or TPU (Tensor Processing Unit) to accelerate model training and efficiently process massive amounts of data.

2) Distributed Training: Apply distributed training to shorten the training time of large-scale models. Technologies like RAY for distributed computing and libraries like Hugging Face's DeepSpeed, which specializes in deep learning training optimization, can be utilized for efficient parallel processing.

3) Accelerating Technologies: Leverage hardware-accelerating technologies to speed up model training. CUDA and cuDNN are libraries that assist in accelerating deep learning models, especially on NVIDIA GPUs.

4) Batch Normalization: Enhance model stability and accelerate training by incorporating

batch normalization. Stabilize the distribution between layers in deep learning models to improve performance (e.g., add batch normalization layers to the model architecture).

5) Data Augmentation: Utilize data augmentation techniques to increase the diversity of text data. Techniques such as shuffling word orders within sentences or inserting synonyms help the model learn from various contexts.

6) Experiment Logging and Monitoring: Use tools like TensorBoard or preferred logging tools to monitor performance during training. Log important metrics during training to monitor the model's performance in real-time and quickly identify issues during the training process.

7) Hyperparameter Tuning: Use hyperparameter optimization tools to automatically find the optimal combination. Techniques like Grid Search or Random Search help adjust hyperparameters, maximizing model performance.

Fine-tuning execution environment setup requires careful planning and supervision to ensure experiment efficiency and model convergence. Regularly optimize based on experiment results within the configured environment to achieve the best possible performance.

### 3.3. Considerations for Fine-Tuning Financial Specialized LLM

Fine-tuning a language model (FLM) for the financial domain involves several steps to ensure the model's effectiveness in understanding and generating financial text. Here are key considerations.

### 3.3.1. Building Domain-Specific Vocabulary Considering Financial Data Characteristics

Financial data possesses unique features, including the impact of stock price fluctuations, sentiment in financial news, and various other aspects. To fine-tune the model effectively, it's crucial to consider these characteristics:

1) Usage of Financial Terminology: Incorporate financial jargon and terminology into the model's vocabulary. Utilize pre-training with financial domain-specific dictionaries or glossaries to enhance the model's understanding of terms like "stock price decline," "exchange rate increase," or "interest rate hike."

2) Handling Numerical Data: Financial data often involves numerical values such as stock indices, currency exchange rates, and interest rates. Employ numerical processing capabilities to ensure accurate handling of numerical information within the text.

3) Complexity of Rules: Financial data follows intricate rules and patterns. Enhance the model's ability to comprehend and apply complex financial rules. For instance, understanding statements like "KOSPI index at 2,300" or "USD exchange rate at 1,300 won".

### 3.3.2. Applying Fine-Tuning Algorithms

Consider fine-tuning algorithms that not only work well with general text but also address the specific characteristics of financial data:

1) LSTM (Long Short-Term Memory): Ideal for handling time-series data, LSTM retains information from the past to process the present. Given the time-sensitive nature of financial data, LSTM can be beneficial for predicting market situations based on historical data.

2) Attention Mechanism: Enables the model to focus on specific portions of input text, allowing it to learn intricate patterns in financial data effectively. Useful for emphasizing the impact of specific events or news in financial predictions.

3) Specialized Financial Algorithms: Algorithms tailored for financial tasks, such as statistical methods for risk management or models like Black-Scholes for option valuation.

Incorporating these algorithms helps the model adapt more effectively to specific financial tasks.

### 3.3.3. Security and Regulatory Compliance

Given the sensitivity of financial data, ensure compliance with security and regulatory standards during fine-tuning. At this stage, the dataset is selected considering data security and personal information protection. When undertaking an LLM fine-tuning in finance, you should consider the following:

1) Data Security: Implement encryption, access controls, and backup measures to secure financial news data.

2) Personal Data Protection: Adhere to regulations regarding the collection, use, and processing of personal information within financial news data.

3) Regulatory Compliance: Comply with financial regulations such as the Electronic Financial Transactions Act or Foreign Exchange Transactions Act when collecting or utilizing financial information.

By addressing these considerations, the fine-tuned LLM can offer high accuracy and utility for specific financial tasks, empowering financial professionals in activities like prediction, research, and report generation with the use of LLM in a secure and compliant manner.

### 3.4. Configuration of LLM for the Financial Specific Domain

In the field of finance, models specialized in natural language processing, such as BloombergGPT and FinGPT, have gained prominence. These models excel in financial tasks by learning financial terms and domain-specific features through fine-tuning on top of general language models.

#### 3.4.1. LLM Pre-trained for Finance from Scratch

Models initially trained for the financial domain include Fin-T5, released in February 2023, and BloombergGPT, introduced in March. Fin-T5 is based on the 770M-T5 model and trained on an extensive dataset of 80 billion finance tokens. BloombergGPT, a GPT-based model specializing in financial language, is pre-trained using Bloomberg's financial data and news. The model, named 50B-BLOOM, is initially based on the 363B Finance tokens and 345B public tokens, undergoing fine-tuning. Predominantly designed for tasks like Named Entity Recognition (NER) and Sentiment Analysis, this model, as reported by Shijie, W., et. al. (2023), demonstrates outstanding performance in financial market trend prediction and news sentiment analysis.

<Table 2> Evaluation Benchmarks of BloombergGPT.

| Suit | Tasks | What does it measure? |
| --- | --- | --- |
| Public Financial Tasks | 5 | Public datasets in the financial domain |
| Bloomberg Financial Tasks | 12 | NER and sentiment analysis tasks |
| Big-bench Hard (Suzgun et al., 2022) | 23 | Reasoning and general NLP tasks |
| Knowledge Assessments | 5 | Testing closed-book information recall |
| Reading Comprehension | 5 | Testing open-book tasks |
| Linguistic Tasks | 9 | Not directly user-facing NLP tasks |

Additionally, models pre-trained on financial news data, such as FinBERT, utilize the BERT architecture to further learn words, syntax, and meanings within financial contexts. Primarily applied to financial research and trading-related problems, FinBERT successfully extracts domain-specific information, contributing to its effective utilization in extracting insights from financial data.

These models showcase the effectiveness of training language models from scratch for the financial domain, providing superior performance in various financial tasks and applications.

### 3.4.2. LLM Fine-Tuned for Financial Specific Domain

Models fine-tuned specifically for the financial sector include FinGPT and Fin-LLaMA, introduced in July 2023. FinGPT, based on OpenAI's GPT architecture, is tailored for finance by using ChatGLM-6B as the base model. It undergoes lightweight fine-tuning with 50,000 samples, employing LoRA technology to learn the intricacies of financial text. FinGPT demonstrates outstanding performance in applications such as financial research, predictive analysis, and automated trading.

Fin-LLaMA, on the other hand, utilizes the LLaMA-33B as its base model and undergoes instruction fine-tuning with 16,900 data samples. Known for its high performance, Fin-LLaMA excels in various financial tasks.

Typically, open-source LLMs like LLaMA2, Falcon, and BLOOM serve as base models, incorporating additional financial-specific information through pre-training or fine-tuning. Each model exhibits unique features, strengths, and weaknesses. FinBERT, for instance, excels in specific tasks but may be sensitive to the quantity of data available. BloombergGPT leverages real financial data and a robust fine-tuning mechanism, making it suitable for practical applications. FinGPT, based on the GPT architecture, achieves high performance through fine-tuning with finance-specific datasets.

These LLM models in the financial domain are expected to provide even higher precision and efficiency in the future, offering substantial value to financial professionals.

### 3.5. Evaluation Criteria and Metrics

Evaluation criteria and metrics are crucial factors for quantitatively and qualitatively assessing the performance of financial-specialized LLMs. To assess the model's quantitative performance, various metrics, such as accuracy, precision, recall, and F1 score, are employed. These metrics provide a quantitative measure of how effectively the model performs specific financial tasks.

<Table 3> Evaluation Criteria and Metrics for Financial-Specialized Language Models

| Evaluation Aspect | Metrics | Criteria |
|---|---|---|
| **Quantitative Performance** | Sentence Generation Accuracy | Sentence Generation Accuracy measures the precision with which generated sentences align with the correct answers. Given that one of the primary objectives of financial-specialized LLMs is to produce accurate and meaningful financial sentences, this metric evaluates the accuracy of the generated sentences. |
| | Financial Prediction Accuracy | Financial Prediction Accuracy assesses the model's precision in predicting outcomes based on given financial data. In the financial domain, where tasks such as stock price prediction or forecasting financial events are crucial, evaluating the model's accuracy in these predictions becomes paramount. |
| | Sentiment Analysis Accuracy | Sentiment Analysis Accuracy evaluates the precision of sentiment analysis results for financial news or research. It measures how effectively the model analyzes the sentiment of financial text, ensuring accurate classification of positive or negative sentiments. |
| | Precision | Sentiment Analysis Accuracy evaluates the precision of sentiment analysis results for financial news or research. It measures how effectively the model analyzes the sentiment of financial text, ensuring accurate classification of positive or negative sentiments. |
| | Recall | Recall computes the ratio of true positive predictions to the total instances that actually belong to the positive class. Particularly important in the financial domain, recall measures how well the model can detect positive instances among the actual positive samples, providing insights into the model's ability to capture relevant information. |
| | F1 Score | F1 Score, calculated as the harmonic mean of precision and recall, offers a balanced metric for evaluating the model's performance. It serves as an indicator of the model's ability to maintain a balance between precision and recall, providing a comprehensive assessment of its overall effectiveness. |
| **Qualitative Performance** | Context Understanding | Context Understanding assesses how well the model comprehends specific contexts within the financial domain. It evaluates the model's ability to grasp domain-specific vocabulary and context nuances in given financial sentences. |
| | Domain-specific Terminology Usage | Domain-specific Terminology Usage measures the model's proficiency in correctly employing specialized financial terms. The accurate utilization of financial terminology indicates how well the model has learned the nuances of the domain. |
| | Text Coherence | Text Coherence evaluates the consistency of generated text by the model. Consistent text is crucial in financial reports and predictive analyses, making it a vital criterion for assessing the model's quality. |
| | Adaptability to Unusual Scenarios | Adaptability to Unusual Scenarios evaluates how effectively the model responds to unexpected situations in the financial domain. Given the dynamic and unpredictable nature of financial markets, the model's adaptability is a crucial aspect of assessment. |
| | Subjective Evaluation by Experts | Subjective Evaluation by Experts involves financial professionals assessing the model's results and providing feedback. Leveraging the experience and knowledge of experts, this evaluation method gauges how practical and effective the model is in specific financial tasks. |
| | Domain-specific Evaluation Metrics | Domain-specific Evaluation Metrics introduce metrics tailored to the important features of the financial domain. By incorporating metrics specialized for financial data and tasks, this evaluation assesses how well the model fits domain-specific requirements. |
| | Text Generation Evaluation | Text Generation Evaluation utilizes metrics such as BLEU Score and ROUGE Score to assess the model's text generation capabilities. This evaluates the model's ability to generate grammatically appropriate and meaningful results, particularly in tasks like financial report generation and news summarization. |

In addition to quantitative evaluation, incorporating subjective assessments from domain experts is essential for a comprehensive evaluation of the model's performance. This involves considering how well the model has learned financial domain knowledge and how effectively it has been applied. Table 3 presents example evaluation criteria and metrics that can be utilized to assess language models specialized for the financial domain across various aspects.

By comprehensively considering these quantitative and qualitative performance metrics, the model's suitability for the financial domain is assessed. The evaluation results are utilized for ongoing improvement and optimization of the model. The evaluation criteria and metrics aid in identifying the strengths and weaknesses of the model, ultimately contributing to the assessment of its utility in the financial domain.

## 4. Fine-tuning Implementation and Utilization Research

In this chapter, we discuss the implementation and application of the Fine-tuning method based on the techniques introduced in Chapter 3. The utilization areas of specialized language models (LLM) in the financial domain are diverse. Through this discussion, we aim to confirm the effective application possibilities of LLM in the financial sector, presenting areas where LLM can enhance decision-making and operational efficiency in the financial domain.

### 4.1. Implementation of LLM Fine-tuning

In this section, we present the key implemented code based on the Fine-tuning procedure illustrated in Figure 8. Python was used as the programming language for implementation, and for tracking parameters such as model weights and biases, the MLOps tool WandB was employed, providing a dashboard. The development infrastructure for training utilized Google Colab, allowing immediate application without additional installations for GPU and Python.

### 4.1.1. Data Collection and Preprocessing

For specialized Fine-tuning in the financial domain, one can construct a proprietary dataset or utilize open datasets. Figure 9 illustrates preprocessed securities and financial terms data in a Question-Answer (QA) set format, prepared as a CSV file (e.g., "##Question: What is an Index?## Answer: An Index measures the performance of a group of stocks serving as a benchmark."). The financial dataset, named 'FinancialStockTerms_Eng' consisting of preprocessed QA pairs, is loaded as the dataset for Fine-tuning, as depicted in Figure 10.

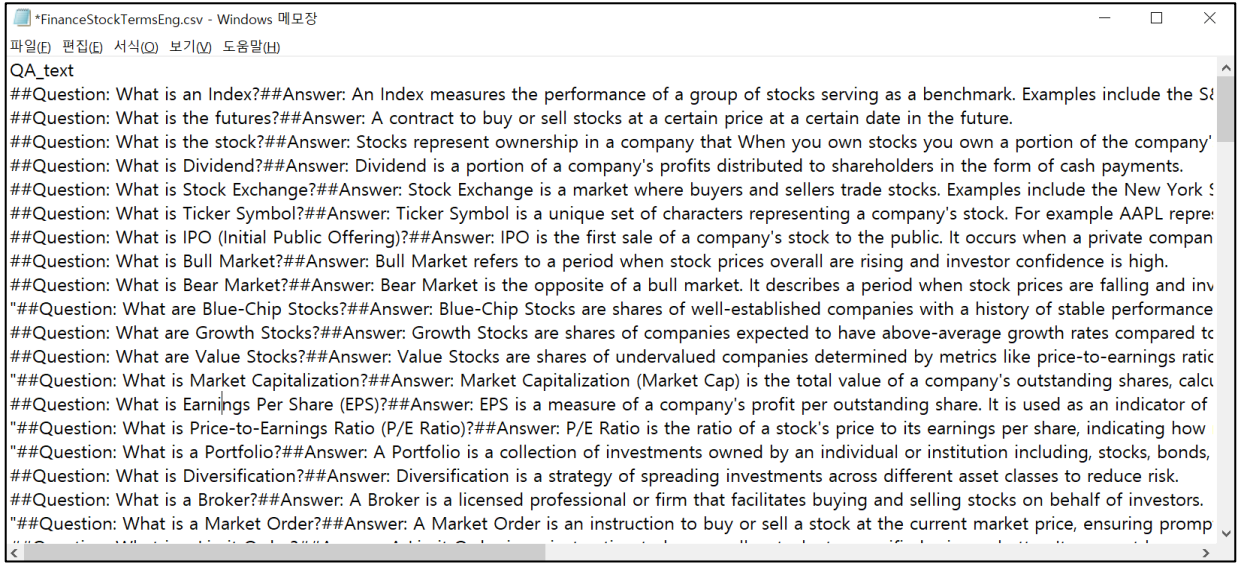

<Figure 9> Financial dataset preparation

```
1 data = load_dataset("csujeong/FinancialStockTerms_Eng")
2
3 data
```

Downloading data: 100% ████████████████████ 26.0k/26.0k [00:00<00:00, 254kB/s]
Generating train split: 135/0 [00:00<00:00, 3675.99 examples/s]
DatasetDict({
    train: Dataset({
        features: ['QA_text'],
        num_rows: 135
    })
})

<Figure 10> Loading dataset for Fine-tuning

### 4.1.2. Selection of Pretrained Language Model (PLM)

In this study, the selection of the pretrained language model (PLM) takes into consideration the model's availability for research and commercialization. To facilitate research and application in various fields, the Falcon 7B model, openly available as open source, was chosen and implemented, as depicted in Figure 11. Quantumization information is stored using BitsAndBytesConfig to enable quantization and is utilized during model loading (AutoModelForCausalLM.from_pretrained).

```
1 model_name = "tiiuae/falcon-7b"  # falcon-7b model
2
3 bnb_config = BitsAndBytesConfig(
4     load_in_4bit=True,           # load model in 4-bit precision
5     bnb_4bit_quant_type="nf4",   # pre-trained model should be quantized in 4-bit NF format
6     bnb_4bit_use_double_quant=True, # Using double quantization as mentioned in QLoRA paper
7     bnb_4bit_compute_dtype=torch.bfloat16, # Pre-trained model should be loaded in BF16 format
8 )
9
10 model = AutoModelForCausalLM.from_pretrained(
11     model_name,
12     quantization_config=bnb_config, # Use bitsandbytes config
13     device_map="auto",  # "auto"so that HF Accelerate will determine which GPU to put each layer of the model on
14     trust_remote_code=True, # Set trust_remote_code=True to use falcon-7b model with custom code
15 )

config.json: 100%                                           1.05k/1.05k [00:00<00:00, 50.4kB/s]
configuration_falcon.py: 100%                               7.16k/7.16k [00:00<00:00, 472kB/s]
```

< Figure 11> PLM Selection

### 4.1.3. Hyperparameter Configuration and Fine-Tuning Training

Once the preparations for quantization are completed, hyperparameters are set as illustrated in Figure 12. For QLoRA-related configurations, LoraConfig is established, and the PEFT model is obtained, as demonstrated in Figure 12.

```
1 model = prepare_model_for_kbit_training(model)
2
3 lora_alpha = 32 # scaling factor for the weight matrices
4 lora_dropout = 0.05 # dropout probability of the LoRA layers
5 lora_rank = 32 # dimension of the low-rank matrices
6
7 peft_config = LoraConfig(
8     lora_alpha=lora_alpha,
9     lora_dropout=lora_dropout,
10    r=lora_rank,
11    bias="none",  # setting to 'none' for only training weight params instead of biases
12    task_type="CAUSAL_LM",
13    target_modules=[           # Setting names of modules in falcon-7b model that we want to apply LoRA to
14        "query_key_value",
15        "dense",
16        "dense_h_to_4h",
17        "dense_4h_to_h",
18    ]
19 )
20
21 peft_model = get_peft_model(model, peft_config)
```

```
1 output_dir = "/content/gdrive/MyDrive/LLM/Falcon-7B-Fintued-Finance-Stock-E"
2 per_device_train_batch_size = 2 # reduce batch size by 2x if out-of-memory error
3 gradient_accumulation_steps = 2  # increase gradient accumulation steps by 2x if batch size is reduced
4 optim = "paged_adamw_32bit" # activates the paging for better memory management
5 save_strategy="steps" # checkpoint save strategy to adopt during training
6 save_steps = 10 # number of updates steps before two checkpoint saves
7 logging_steps = 10  # number of update steps between two logs if logging_strategy="steps"
8 learning_rate = 2e-4  # learning rate for AdamW optimizer
9 max_grad_norm = 0.3 # maximum gradient norm (for gradient clipping)
10 max_steps = 60        # training will happen for 60 steps
11 warmup_ratio = 0.03 # number of steps used for a linear warmup from 0 to learning_rate
12 lr_scheduler_type = "cosine"  # learning rate scheduler
```

< Figure 12> Hyperparameter Configuration

The final step involves the sequence of model training. Figure 13 showcases the code and results for fine-tuning execution, which involves training the PLM with additional datasets. The resulting Fine-Tuned Language Model (FLM) can be effectively utilized for domain-specific tasks.

< Figure 13> Fine-Tuning Training Execution

4.1.4. FLM Verification

To test the generated FLM, the PEFT FLM is loaded, as depicted in Figure 14. This step ensures that the model operates correctly.

< Figure 14> PEFT Model Loading

< Figure 15> PEFT Model Testing

When posing the question 'What is the Index?' to both the pretrained PLM and the PEFT fine-tuned PLM, differences in responses are observed. The pre-fine-tuned PLM generates a generic answer unrelated to stocks (e.g., "A list of all the pages in a website."), whereas the FLM, fine-tuned with the specialized securities and finance dataset through PEFT, produces an answer related to stocks (e.g., "An index is a measure of the performance of a group of stocks or the overall market.")

In summary, we have explored the method of fine-tuning PLMs with domain-specific datasets, validated the results, and demonstrated how FLMs can be tailored for specific tasks. Depending on the domain classification and the scale of business, fine-tuning various LLMs based on specific needs can result in customized FLMs suitable for different applications.

4.2. Application Fields and Cases of Financial LLM

4.2.1. Financial Prediction and Trading

In the realm of financial prediction and trading, leveraging LLM facilitates predicting market trends through activities like stock price prediction and sentiment analysis of financial news. For instance, models can analyze financial news, corporate reports, and market trends collectively to forecast future stock movements. Investors can then make more informed decisions based on these predictions. Sentiment analysis of financial news provides insights into the emotions of market participants. LLMs analyze sentiments in financial news, offering investors information on market trends based on positive or negative news, enabling strategic trading decisions. A notable example is Mirae Asset Securities, which has introduced a service utilizing ChatGPT to summarize stock market conditions (Korea Financial Times, 2023).

### 4.2.2. Automated Financial Document Processing

In the domain of automated financial document processing, efficiency gains can be achieved through activities like contract analysis and assistance in financial report generation. Contract analysis involves utilizing LLMs to automatically analyze and extract essential information from various contracts, enhancing efficiency and reducing errors in financial workflows. Additionally, LLMs can aid in financial report generation, summarizing information and assisting financial experts in swiftly producing reports. Real-world examples include Samsung Life's automation of insurance claims payment processes using AI-based Optical Character Recognition (OCR) (AITimes, 2023) and JP Morgan's 'COiN,' employing AI to analyze corporate loan contracts, classify types, and extract key phrases, reducing analysis time and improving accuracy (Financial Focus, 2023).

### 4.2.3. Financial Research and Information Extraction

In the realm of financial research and information extraction, LLMs prove valuable in extracting useful information from financial datasets for research purposes or recommending financial products aligned with customer investment preferences. LLMs efficiently extract information necessary for financial research by collecting and synthesizing financial data from various sources, enabling financial experts to access the latest information for more effective decision-making. Examples include Korea Investment & Securities' AI-based research service, 'AIR Listing Index Fund (ETF),' and DB Insurance's use of AI to analyze relational data, detect insurance fraud, and improve accuracy (Korea Financial Times, 2023; Data Hunt, 2023).

### 4.2.4. Customer Interaction and Service Enhancement

In the domain of customer interaction and service enhancement, automatic response systems and personalized product recommendations contribute to increased customer satisfaction. Automatic response systems, powered by LLMs, efficiently address diverse customer inquiries, reducing response times and enhancing efficiency. LLMs can also analyze customer financial histories and preferences to recommend tailored financial products, increasing customer satisfaction and improving service quality. Examples include KB Kookmin Card and KBpay's 'Event Q&AI' service, providing marketing event information through natural language dialogue (AITimes, 2023), Toss's 'Ask GPT' feature for conversational interactions in their app (AITimes, 2023), and NongHyup Bank's 'ARMI AI,' utilizing AI chatbots for automating customer satisfaction surveys and extracting statistics and analysis results automatically (Financial Focus, 2023).

These applications demonstrate that integrating LLMs into the financial domain automates various tasks, supports decision-making, and can provide a competitive edge for financial institutions, enhancing adaptability in the rapidly changing financial market.

## 5. Discussion and Conclusion

This study has presented the fine-tuning of Language Models (LLMs) specialized in the financial domain and explored various applications. By considering the characteristics of financial data, fine-tuning was conducted to enhance the model's performance. The study delved into applying LLMs to diverse tasks in finance, such as financial prediction, automation, research, and customer interaction. Notably, the step-by-step validation results were provided for the procedures of financial domain dataset selection, data preprocessing, pre-trained LLM model selection, hyperparameter tuning, fine-tuning execution environment setup, evaluation metrics for fine-tuning performance, and model generation.

The study highlighted considerations for fine-tuning in the financial domain, examining areas such as prediction accuracy improvement, workflow efficiency enhancement through automation, and the facilitation of research and customer service. In the context of financial prediction, an enhanced prediction accuracy aids investors in making more precise decisions, while automation contributes to efficiency gains, reducing processing times in tasks like contract analysis and financial report generation. Additionally, LLMs can be instrumental in research, efficiently extracting information for research reports, thereby supporting decision-making and strategic planning within financial institutions. Regarding customer interaction improvement, automated response systems and personalized product recommendations can enhance the quality of customer service.

However, the study focuses primarily on the methodology of fine-tuning model creation, leading to certain limitations. The dataset's limited diversity may result in the model being overly biased toward specific domains, necessitating a more extensive and representative dataset. Additionally, overcoming the model's generalization limitations may require further parameter tuning and optimization of the deep network structure. The absence of financial domain knowledge in the model could be addressed by exploring more effective ways to convey domain-specific knowledge.

Future research directions could involve expanding the study using more diverse and representative financial datasets to strengthen the model's learning. Alongside optimizing the model structure, reinforcing financial domain knowledge could further enhance the model's performance. Moreover, exploring the broader application of LLMs in various financial tasks such as bankruptcy prediction, investment portfolio optimization, credit scoring, and extending research to other domains like manufacturing and public services is essential.

Finally, ethical considerations not covered in this study deserve attention. Research in the financial sector involving sensitive information should emphasize aspects such as personal data protection and fair decision-making to enhance the model's trustworthiness.

In conclusion, this study has explored the possibilities and limitations of applying LLMs to the financial domain, providing a foundation for specialized domain research. The significance and value lie in actively utilizing LLMs in financial services within enterprises, paving the way for future advancements in this specialized field.

# References

[1] AITimes. (2023, October 25). Samsung Life, "Innovation in the Insurance Industry Document Automation!"... Achieving Document Automation with 'DocAI' AI OCR Solution on the Upstage without Human Intervention. https://www.aitimes.kr/news/articleView.html?idxno=29203

[2] AITimes. (2023, October 25). KB Kookmin Card Launches 'Event Q&AI' Beta Service for Customer Experience Innovation Based on LLM. Link

[3] Bommasani, R., et. al. (2022, July 12). On the Opportunities and Risks of Foundation Models. https://arxiv.org/pdf/2108.07258.pdf

[4] Chowdhery, A., et., al. (2022). Palm: Scaling language modeling with pathways. CoRR, vol. abs/2204.02311

[5] Data Hunt. (2023, October 19). Insurance AI Artificial Intelligence, Applied Technologies, and Use Cases. https://www.thedatahunt.com/trend-insight/ai-in-insurance

[6] Dettmers, T., et. al. (2023, May 23). QLoRA: Efficient Fine-tuning of Quantized LLMs, https://arxiv.org/pdf/2305.14314.pdf

[7] Dilmegani, C. (2023, June 21). Large Language Models: Complete Guide in 2023. https://research.aimultiple.com/large-language-models/

[8] Edward, H., et. al. (2021, October 16). LORA: LOW-RANKADAPTATION OF LARGE LANGUAGEMODELS, https://arxiv.org/pdf/2106.09685.pdf

[9] Financial Focus. (2023, October 31). Birth and Evolution of ChatGPT, Generative AI, Contributes to Increased Productivity in Banking Business. http://ffnews.co.kr/detail.php?number=4768&thread=28r38

[10] Jeong, C. S., & Jeong, J. H. (2020). A Study on the Method of Implementing an AI Chatbot to Respond to the POST COVID-19 Untact Era, *Journal of Information Technology Services*, 19(4), 31–47. https://doi.org/10.9716/KITS.2020.19.4.031

[11] Jeong, C. S. (2023a). A Study on the RPA Interface Method for Hybrid AI Chatbot Implementation, *KIPS Transactions on Software and Data Engineering*, 12(1), 41-50. https://doi.org/10.3745/KTSDE.2023.12.1.41

[12] Jeong, C. S. (2023b). A Case Study in Applying Hyperautomation Platform for E2E Business Process Automation, *Information Systems Review*, 25(2), 31-56. https://doi.org/10.14329/isr.2023.25.2.031

[13] Jeong, C. S. (2023c). A Study on the Service Integration of Traditional Chatbot and ChatGPT, *Journal of Information Technology Applications & Management*, 3(4), 11-28. https://doi.org/10.21219/jitam.2023.30.4.001

[14] Jeong, C. S. (2023d). A Study on the Implementation of Generative AI Services Using an Enterprise Data-Based LLM Application Architecture. *Advances in Artificial Intelligence and Machine Learning*, 3(4). 1588-1618. https://dx.doi.org/10.54364/AAIML.2023.1191

[15] Jeong, C. S. (2023e). Generative AI service implementation using LLM application architecture: based on RAG model and LangChain framework, *Journal of Intelligence and Information Systems*, 29(4), 129-164. https://dx.doi.org/10.13088/jiis.2023.29.4.129

[16] Jeong, J. H. and Jeong, C. S. (2022). Ethical Issues with Artificial Intelligence (A Case Study on AI Chatbot & Self-Driving Car), *International Journal of Scientific & Engineering Research*, 13(1). 468–471.

[17] Korea Financial Times. (2023, May 2). Securities Industry Riding the AI Wave with ChatGPT... "Steady in Research and Investment." https://www.fntimes.com/html/view.php?ud=20230429032803830dd55077bc2_18

[18] Mayank, S. (2023, June 30). Generative AI: Empowering Innovation with its Astonishing Capabilities. https://shurutech.com/innovating-with-generative-ai/

[19] Nijkamp, E., et., al. (2022). Codegen: An open large language model for code with mtulti-turn program synthesis. arXiv preprint arXiv:2203.13474

[20] Raschka, S. (2023, May 20). Fine-tuning LLMs Efficiently with Adapters. https://magazine.sebastianraschka.com/p/Fine-tuning-llms-with-adapters

[21] Shijie, W., et. al. (2023, May 9). BloombergGPT: A Large Language Model for Finance, https://arxiv.org/pdf/2303.17564.pdf

[22] Taylor, R., et., al. (2022). On the Opportunities and Risks of Foundation Models. CoRR, vol. abs/2211.09085

[23] Zhao, W. X., et al. (2023, June 29). A Survey of Large Language Models. https://arxiv.org/pdf/2303.18223.pdf